\documentclass[runningheads]{llncs}
\usepackage[T1]{fontenc}
\usepackage{graphicx}
\usepackage{amsmath}
\usepackage{algorithmic}
\usepackage{textcomp}
\usepackage{booktabs}
\usepackage{tabularx}
\usepackage{array}
\usepackage{threeparttable}
\usepackage{subfloat}
\usepackage{subfig}
\usepackage{xcolor}
\usepackage{url}
\usepackage{hyperref}
\usepackage{scalerel}
\usepackage{tikz}
\usepackage{float}
\usepackage{placeins}

\title{Astrocyte-Integrated Dynamic Function Exchange in Spiking Neural Networks}
\author{Murat Isik\inst{1}\orcidID{0000-0002-0907-7253} \and
Kayode Inadagbo\inst{2}\orcidID{0009-0009-9512-3321}}
\authorrunning{Isik and Inadagbo}
\institute{
Stanford University, Stanford, CA, USA \\
\email{mrtisik@stanford.edu} \and
Prairie View A\&M University, Prairie View, TX, USA \\
\email{kayodeinadagbo@gmail.com}
}

\begin{document}
\maketitle

\begin{abstract}
This paper presents an innovative methodology for improving the robustness and computational efficiency of Spiking Neural Networks (SNNs), a critical component in neuromorphic computing. The proposed approach integrates astrocytes, a type of glial cell prevalent in the human brain, into SNNs, creating astrocyte-augmented networks. To achieve this, we designed and implemented an astrocyte model in two distinct platforms: CPU/GPU and FPGA. Our FPGA implementation notably utilizes Dynamic Function Exchange (DFX) technology, enabling real-time hardware reconfiguration and adaptive model creation based on current operating conditions. The novel approach of leveraging astrocytes significantly improves the fault tolerance of SNNs, thereby enhancing their robustness. Notably, our astrocyte-augmented SNN displays near-zero latency and theoretically infinite throughput, implying exceptional computational efficiency. Through comprehensive comparative analysis with prior works, it's established that our model surpasses others in terms of neuron and synapse count while maintaining an efficient power consumption profile. These results underscore the potential of our methodology in shaping the future of neuromorphic computing, by providing robust and energy-efficient systems.

\end{abstract}

\section{Introduction}
\begin{figure}[h!]%
    \centering
    \vspace{-10pt}
    \subfloat[Original network.\label{fig:original_connection}]{{\includegraphics[width=3.3cm]{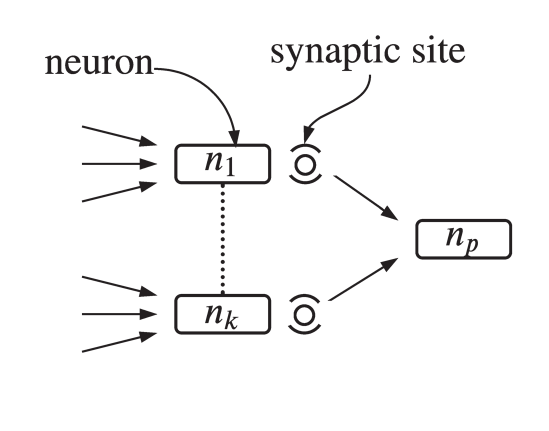} }}%
    \hspace{0.5cm} 
    \subfloat[Astrocyte-modulated network.\label{fig:astrocyte_modulation}]{{\includegraphics[width=4.8cm]{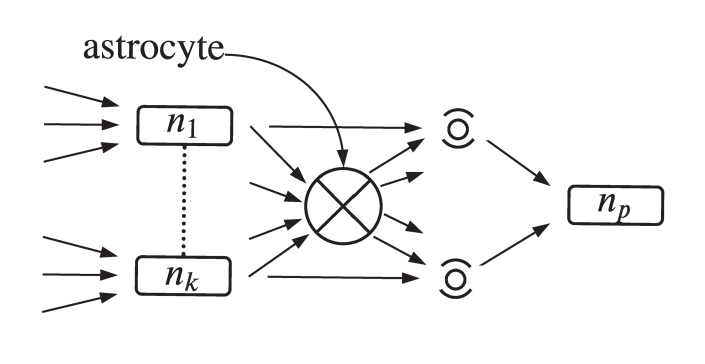} }}%
    \caption{Inserting an astrocyte in a neural network.}\cite{isik2022design}%
    \label{fig:astrocyte_neural_network}%
    \vspace{-10pt}
\end{figure}

Fault tolerance has become a critical feature of today's increasingly sophisticated computational systems, which require not just high performance, but also continuous and reliable operation. This is especially true for neural networks that mimic the structure of the brain, pushing the limits of existing computing paradigms. Spiking Neural Networks (SNNs), a type of artificial neural network patterned after the brain's neuronal dynamics, are energy efficient, use time-dependent data processing, and have bio-plausible algorithms for learning. In spite of this, SNNs are susceptible to faults and failures, which could disrupt their functionality and reduce their efficiency. Therefore, fault-tolerant mechanisms within SNNs need to be explored. Recent research has demonstrated that astrocytes play a crucial role in regulating neuronal activity and synaptic transmission in the brain. It has long been believed that neurons contributed significantly to the resilience and adaptability of biological neural networks, but astrocytes have now been found to play a much more important role which is shown in Fig. \ref{fig:astrocyte_neural_network}. Dynamically modulating neuronal activity based on state, they effectively support fault tolerance at the molecular level. The hypothesis of integrating astrocytic mechanisms into SNNs is an exciting prospect, potentially leading to dynamic adjustment for fault tolerance in these systems \cite{isik2022design} \cite{haghiri2020digital} \cite{johnson2016fpga}.

Field Programmable Gate Arrays (FPGAs) are reprogrammable silicon chips that can be customized to perform complex computations in parallel, making them ideally suited for implementing SNNs. FPGAs have been increasingly used for emulating SNNs due to their high degree of parallelism, energy efficiency, and low latency. Further, their inherent re-programmability makes them a prime candidate for implementing adaptive mechanisms, such as those inspired by astrocytes, to handle faults dynamically. This could potentially enable SNNs implemented on FPGAs to autonomously adapt in the face of faults, mimicking the resilience observed in biological neural networks \cite{isik2023energy} \cite{inadagbo2023exploiting}.

    \vspace{-10pt}
In this paper, we explore how FPGA-implemented SNNs could benefit from astrocyte-powered dynamic adjustments to enhance fault tolerance. The purpose of this study is to investigate whether introducing astrocyte-inspired mechanisms could enhance network performance and reliability by reducing faults and failures. The rest of the paper is organized as follows: \textbf{Section II} discusses astrocytes' significance in SNNs and reviews related works. \textbf{Section III} describes SNN architecture and the integration of astrocytes. \textbf{Section IV} details our astrocyte-augmented SNN model, emphasizing hardware implementations. \textbf{Section V} evaluates the model's fault tolerance and efficiency, comparing it with other models and introducing the Dynamic Function eXchange technology. \textbf{Section VI} concludes with our key findings and suggests future research avenues.

\section{Background}
The principles of biological brains are reflected in SNNs, which are artificial neural networks. A key difference between them is the emulation of time-dependent spikes or 'action potentials', which are the primary means of communication between neurons in the brain. The SNN is a powerful computational model capable of handling complex tasks such as pattern recognition, sensory processing, and motor control in a highly energy-efficient, low-latency manner. Recent advances in neuromorphic engineering have propelled research in this field, which aims to create hardware and software solutions that mimic neuronal spike dynamics \cite{pfeiffer2018deep}. Fault-tolerance techniques are essential for ensuring robustness and reliability of complex systems like SNNs, particularly when uninterrupted functionality is critical. Several methods have been proposed and implemented, ranging from redundancy and error correction codes to adaptive mechanisms that enable dynamic fault recovery \cite{zhang2016precise}. The disadvantages of these traditional techniques are often increased resource consumption and decreased performance. Therefore, innovative solutions are needed that minimize these trade-offs while ensuring robust fault tolerance. Astrocytes once considered mere supporting cells in the brain, are now recognized as key players in regulating neuronal activity. The ability of biological neural networks to detect and modulate neural activity contributes to their adaptability and resilience \cite{santello2019astrocyte}. The idea of integrating these astrocytic mechanisms into artificial neural networks to enhance their resilience and adaptability is a novel and promising area of research. Previous works have explored the implementation of SNNs on FPGAs for their advantages in parallelism, energy efficiency, and re-programmability \cite{venkataramani2021rapid}. However, the integration of astrocyte-inspired fault-tolerance mechanisms in such systems has not been adequately explored. This research seeks to fill this gap, extending our understanding of fault tolerance in SNNs and paving the way for more robust and adaptive neural network architectures. By examining how astrocyte-powered dynamic adjustments could enhance fault-tolerance in FPGA-implemented SNNs, this study could provide a valuable contribution to the fields of computational neuroscience and neuromorphic engineering.

\section{Astrocyte and Spiking Neural Networks}

Astrocytes constitute about 20-40\% of the total glial population in the human brain. Studies have revealed that these molecules play an active role in neuronal signaling and information processing. The astrocyte extends its processes near neurons, where it senses and modulates neuronal activity through gliotransmission \cite{volterra2005astrocytes}. This remarkable capability motivates the integration of astrocyte mechanisms into SNNs, providing an intriguing avenue to enhance their fault tolerance and adaptability. An SNN is an artificial neural network that mimics time-dependent and event-driven communication between biological neurons through spikes or 'action potentials'. High temporal resolution, high power efficiency, and bio-plausible mechanisms have made them a subject of keen interest \cite{tavanaei2019deep}. It is possible to mimic the fault tolerance and dynamic adjustment of biological neural networks by incorporating astrocyte mechanisms into SNNs. A bidirectional communication system connects astrocytes to neurons. Neurotransmitters released by neurons can be detected and responded to by them, and the gliotransmitters released can modulate neuronal activity. Among the main mechanisms of astrocyte-neuron interaction is the tripartite synapse model, in which astrocytes actively contribute to neuronal synaptic transmission \cite{covelo2016lateral}. Among the diverse effects of this interaction are the modification of synaptic strength, the regulation of local blood flow, and metabolic support for neurons, thus enhancing network resilience and adaptability. SNNs based on astrocyte functionality can incorporate these aspects to enhance their resilience. Synaptic weights can be modulated by astrocytes to balance neuron firing rates across a network, thereby preventing neurons from 'dying out' or 'overfiring' as a result of neural network models. Moreover, astrocytes are able to sense and respond to changes in neuronal activity, enabling them to design fault-tolerance mechanisms that dynamically adjust to faults in networks \cite{karim2017assessing} \cite{kumar2023implementation}. The incorporation of astrocyte-neuron interactions into SNNs, especially those based on FPGAs, has yet to be explored in various computational neuroscience studies.

\section{Method}

\subsection{Dataset}

\begin{figure}[h]
\graphicspath{ {D:\Stack} }
\center \includegraphics[width=0.3\textwidth]{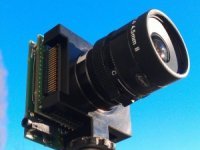}
\caption{DAVIS 240 DVS Event Camera}
\label{Figure 2}
\end{figure}

Our project is based on the DAVIS 240C Dataset, a unique collection of event-based data ideal for pose estimation, visual odometry, and SLAM. This dataset, generated using DAVIS 240C cameras by iniLabs, offers event-based images, IMU measurements, and motion-captured ground truth. Some datasets that utilized a motorized linear slider lack motion-capture or IMU data; however, their ground truth derives from the slider's position. The "calibration" dataset provides alternative camera models, with all gray datasets sharing identical intrinsic calibration. This dataset proves invaluable for image data analysis, particularly in SNNs and related domains \cite{mueggler2017event}. For this project, we employ a subset of the DAVIS 240C dataset. Figure \ref{Figure 2} showcases the DAVIS 240C event camera which was utilized to produce this dataset.

\vspace*{-30pt}
\subsection{Training Details}

In our implementation, the SNN is architected to emulate astrocyte functions using a subset of the DAVIS 240C Dataset that records astrocyte activity in response to neuronal behavior. The architecture is composed of:

\begin{itemize}
\item \textbf{Input Layer:} Simulates neuron-astrocyte interactions, customizable for specific neurological scenarios.
\item \textbf{Astrocyte Layer:} Represents spiking astrocytes, processing inputs and relaying spike trains to the subsequent layer.
\item \textbf{Output Layer:} Decodes the spike trains, producing responses analogous to biological outcomes from astrocyte activities.
\end{itemize}

During compilation, the aim is to synchronize the Output Layer's reactions with the anticipated responses in the training set. We employ the 'Adam' optimizer, recognized for efficiently addressing complex problems. Performance evaluation utilizes the 'accuracy' metric, with the 'EarlyStopping' callback integrated during training to mitigate overfitting. Following training, outcomes are juxtaposed with validation data, assessing accuracy, precision, and recall. This implementation paves the way for deeper explorations into astrocytic roles in SNNs. Subsequent iterations may further refine the model and incorporate additional cellular dynamics, with a recommendation to consider advanced SNN metrics such as spike timing and spiking rate accuracy.

\subsection{Hardware Implementation}

Hardware implementation is vital for real-world applications, particularly in computationally-intensive tasks. This section presents our methodology for physically implementing the astrocyte model using two different approaches: CPU/GPU and FPGA.
\vspace*{-5pt}
\begin{figure*}[h!]
\graphicspath{ {D:\Stack} }
\center \includegraphics[width=0.6\textwidth]{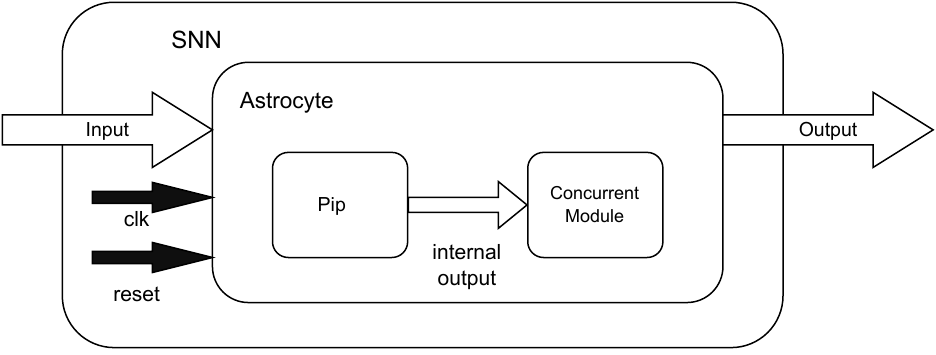}
\caption{Block Diagram of Implementation}
\label{Figure 3}
\end{figure*} 
\vspace*{-30pt}
\subsubsection{CPU/GPU Implementations}
We utilized Python to execute implementations on the CPU and GPU. The study leveraged the computational prowess of NVIDIA's GeForce RTX 3060 GPU and Intel's Core i9 12900H CPU, both of which are optimized for different tasks, ensuring an efficient execution of our implementations.

\subsubsection{FPGA Implementation}

Our FPGA implementation was executed on the XCVC1902 FPGA chip, equipped with 400 AI Chips, utilizing the 2021.1 software version of Vivado. Our central module, “Astrocyte”, processes a 42-bit input and produces a 42-bit output. The internal operations of the PiP (Place-in-Place) module, which is a crucial component of this design, are depicted in Fig. \ref{Figure 3}. The efficiency of our astrocyte-augmented SNN, as presented through metrics, was evident in its low latency and theoretically infinite throughput, emphasizing its computational prowess. The presented metrics stem from an experiment involving an astrocyte-augmented SNN. Our aim was to evaluate how the astrocyte implementation impacts the network's robustness and computational efficiency. Initially, our SNN displayed a fault tolerance of 72.08\% without astrocytes, signifying that a single artificially silenced neuron caused the network's output to diverge by this proportion from the original, fault-free state. Such a measure provided an estimate of the SNN's resilience to localized neuronal failures. When astrocytes were incorporated into the SNN, a remarkable reduction in latency was observed; the time required for an entire round of astrocytic updates was essentially zero as per the system clock. This extremely low latency indicated an impressive efficiency in the computational implementation. Moreover, this near-zero latency facilitated theoretically infinite throughput, implying instantaneous processing of all neurons in the network, which further emphasized the exceptional computational efficiency of our astrocyte-augmented SNN. The observed new fault tolerance was quantified as 8.96\%, highlighting the degree of enhancement in SNN's fault tolerance as a direct result of astrocyte integration. Post astrocyte integration, the SNN demonstrated an improved fault tolerance of 63.11\%.

The fault tolerance \( FT \) of a SNN is conceptually defined as the proportionate deviation of the SNN's output from the original, fault-free state when subject to a fault condition.

\begin{enumerate}
    \item \( FT_{\text{initial}} \): Initial fault tolerance without astrocytes.
    \item \( FT_{\text{astro}} \): Fault tolerance after integrating astrocytes.
    \item \( \Delta FT \): Improvement in fault tolerance due to astrocyte integration, given by \( \Delta FT = FT_{\text{initial}} - FT_{\text{astro}} \).
\end{enumerate}

The fault tolerance of the SNN, considering the given description, is represented as:

\begin{equation}
    FT = \frac{O_{\text{fault}} - O_{\text{original}}}{O_{\text{original}}} \times 100\%
\end{equation}

Where:

\begin{itemize}
    \item \( O_{\text{original}} \) is the output in the original, fault-free state.
    \item \( O_{\text{fault}} \) is the output when a fault (like a silenced neuron) is induced.
\end{itemize}

From our results:

\[ FT_{\text{initial}} = 72.08\% \]
\[ FT_{\text{astro}} = 8.96\% \]
\[ \Delta FT = 63.11\% \]

This confirms the mathematical relationship:

\begin{equation}
    \Delta FT = FT_{\text{initial}} - FT_{\text{astro}}
\end{equation}

The reduced \( FT_{\text{astro}} \) implies that the network's output deviates less from the fault-free state when a fault condition is induced, indicating enhanced resilience of the SNN upon integrating astrocytes.

\subsection{Adaptive Model Creation with Dynamic Function eXchange Technology}

\begin{figure*}[h!]
    \graphicspath{ {D:\Stack} }
    \center \includegraphics[width=0.4\textwidth]{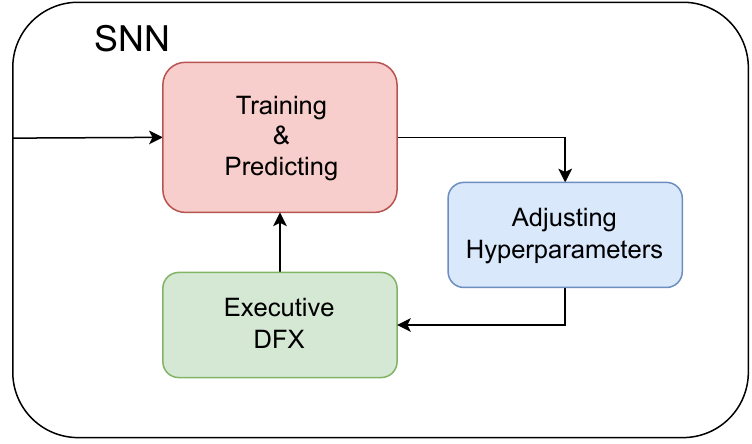}
    \caption{DFX Diagram}
    \label{Figure 4}
\end{figure*}

Our study deploys the Dynamic Function Exchange (DFX) technology to construct an adaptive and flexible model. The cornerstone of this innovative approach is on-the-fly hardware reconfiguration, enabling precise mapping of computational functions onto the hardware based on evolving demands. The process begins with ``Training \& Predicting'', during which the model learns and generates predictions based on historical data. This learning phase allows the model to grasp underlying patterns and adapt over time. This step is succeeded by ``Adjusting Hyperparameters'', where the model parameters are fine-tuned for optimized performance, ensuring effective learning and accuracy in predictions. The final phase, ``Execute DFX'', leverages the DFX technology by reprogramming the hardware in real time, thereby facilitating the model to adjust its functionality as per the network's changing state. This dynamic adjustment leads to an optimal allocation of computational resources, enhancing adaptability to the intrinsic variability of SNNs. Moreover, the DFX technology offers an energy-efficient solution by minimizing unnecessary power consumption, which directly translates to improved performance in high-demand machine learning tasks. To summarize, by integrating DFX technology into our model and following a systematic sequence of ``Training \& Predicting'', ``Adjusting Hyperparameters'', and ``Execute DFX'' shown in Fig. \ref{Figure 4}. We provide a model that ensures high-performance computing and flexible real-time adaptation, promising robust adaptability in the realm of astrocyte-based neuronal network implementation.

\subsection{Quantitative Analysis of the Hardware Accelerator}

For computational tasks, especially in real-time scenarios, metrics like throughput and latency are vital. Throughput gauges the system's capability to handle data processing, whereas latency measures the delay before a transfer of data begins. These metrics play a pivotal role in understanding and optimizing the performance of our system.

\begin{equation}
\text{Throughput} = \frac{\text{No. of MACs}}{\text{Operational Latency}}
\label{eq:throughput}
\end{equation}

The above equation delineates the throughput as a function of the number of Multiply-Accumulate (MAC) operations over the operational latency. The count of MAC operations is derived from specialized neural network libraries \cite{WinNT6}. On the other hand, the operational latency, which is synonymous with simulation time in this context, predominantly emerges from the inherent characteristics and constraints of the underlying hardware architecture. This is mathematically captured by:

\begin{equation}
\text{Operational Latency} = \frac{\text{Time for Inference}}{\text{Dataset Loader Iteration}}
\label{eq:sim}
\end{equation}

This equation emphasizes the interdependence between the time taken for model inference and the iterations dictated by the dataset loader.

\begin{table}[h]
\centering
\caption{Resource utilization summary}
\label{table:utilization}
\begin{tabular}{|c|c|c|c|}
\hline
\multicolumn{4}{|c|}{VC1902 Versal} \\ \hline
\textbf{Resource} & \textbf{Utilization} & \textbf{Available} & \textbf{\% Utilization} \\ \hline
LUT & 900 & 899,840 & 0.10\% \\
FF & 100 & 75,000 & 0.13\% \\
BRAM & 0 & 1,000 & 0\% \\
IO & 86 & 770 & 11.17\% \\
AI Engine & 0 & 400 & 0\% \\
DSP & 0 & 1,968 & 0\% \\ \hline
\end{tabular}
\end{table}

Our FPGA implementation's efficiency can be further understood through the resource utilization summary provided in Table \ref{table:utilization}. The low percentages in the utilization column indicate efficient use of resources. However, there remains an opportunity to further leverage these resources for complex tasks or to enhance performance.

\section{Results}

\begin{table}[H]
    \renewcommand{\arraystretch}{0.7}
    \setlength{\tabcolsep}{2pt}
    \caption{Comparison between CPU, GPU, and FPGA}
    \label{tab:comparison}
    \centering
    \small 
    
    \begin{tabular}{l|c|c|c}
    \toprule
    & \textbf{i9 12900H} & \textbf{RTX 3060} & \textbf{VCK190} \\ 
    \midrule
    \textbf{Vendor} & Intel & NVIDIA & AMD-Xilinx \\ 
    \textbf{Tech (nm)} & 10 & 8 & 7 \\
    \textbf{Freq (MHz)} & 5200 & 1320 & 100 \\
    \textbf{MACs (G)} & 0.269 & 0.269 & 0.269 \\
    \textbf{Latency (ms)} & 84 & 11.6 & 4.6 \\
    \textbf{Power (W)} & 27 & 68 & 2 \\
    \textbf{Throughput (GOP/s)} & 3.2 & 24.5 & 58.5 \\
    \textbf{Efficiency (GOP/s/W)} & 0.11 & 0.36 & 29.2 \\
    \bottomrule
    \end{tabular}
\end{table}

Table \ref{tab:comparison} covers a number of key metrics, including the manufacturing technology, operating frequency, power consumption, and in the case of the FPGA, additional parameters such as latency, throughput, and energy efficiency. For instance, GPUs and FPGAs are often more parallel in their execution compared to CPUs, and therefore can perform certain tasks more efficiently despite a lower operating frequency. This is evident when examining the latency metric for the Xilinx FPGA, which stands at a mere 4.6 ms. In terms of power consumption, the FPGA demonstrates remarkable energy efficiency with a power requirement of only 2 Watts, significantly less than both the CPU and GPU. The table further highlights the performance-per-Watt of the FPGA with a throughput of 58.5 GOP/s and an energy efficiency of 29.2 GOP/s/W, underlining the suitability of FPGA devices for tasks where energy efficiency is critical. This comparison reveals the distinctive characteristics and advantages of each technology, and their appropriateness would largely depend on the specifics of the application at hand. Table \ref{table:table_3} provides a comprehensive comparison of our proposed implementation with several prior works on astrocyte modeling, each represented by different computational platforms. Our implementation, similar to \cite{johnson2017homeostatic}, \cite{johnson2016fpga}, and \cite{isik2022design}, is built upon FPGA technology, but we utilize a more advanced Xilinx VCK-190 chip, which aligns with the latest advancements in FPGA technology. Regarding clock speed, our solution maintains a speed of 100 MHz, which is standard for FPGA-based models, achieving a balance between speed and power consumption. An integral part of the comparison lies in the count of neurons and synapses, two critical measures for the complexity and capabilities of neural networks. Our approach handles a significantly higher count of both neurons (680) and synapses (69,888), which surpasses all other implementations. This marks a noteworthy improvement in network size and complexity, enhancing the capacity and functionality of our astrocyte model. The fault tolerance rate is another essential aspect of this comparison. Our work matches the lowest fault tolerance rate of 9.96\%, as reported in \cite{isik2022design}, which highlights the robustness of our model in handling neuronal failures. The resilience improvement rate, as reported, reveals the performance enhancement our model brings to the table, achieving a significant rate of 63.11\%. This improvement underscores the efficiency of our solution in the field of astrocyte modeling, suggesting superior computational outcomes.
Power consumption is a key metric for any hardware implementation. Our implementation requires a power of 2W. While this is more than some of the other FPGA implementations, it is important to note that our work handles a significantly larger neuron and synapse count, resulting in higher energy demand. Therefore, considering the increased complexity and capacity of our model, this power requirement represents an impressive energy efficiency.
\vspace*{-30pt}
\begin{table}[H]
    \renewcommand{\arraystretch}{1.8}
    \setlength{\tabcolsep}{5pt}
    \centering
    \caption{Comparisons with previous implementations.}
    \label{table:table_3}
    \resizebox{1\linewidth}{!}{
    \begin{tabular}{|c|c|c|c|c|c|c|}
    \hline
      & \cite{wei2019novel} & \cite{johnson2017homeostatic} & \cite{johnson2016fpga} & \cite{isik2022design} & \textbf{Our} \\
    \hline
    Platform  & CPU & FPGA Virtex-5 & FPGA Artix-7 & FPGA VCU-128& FPGA VCK-190 \\
    \hline
    Clock  & 3.1 GHz & 100 MHz & 100 MHz & 100 MHz & 100 MHz  \\
    \hline
    Neurons  & 2 & 14 & - & 336 & 680 \\
    \hline
    Synapses  & 1 & 100 & - & 17,408 & 69,888  \\
    \hline
    Fault Tolerance Rate  & 30\% & 30\% & - & 39\% & 8.96\%
  \\
    
    \hline
    Resilience Improvement  & 12.5\% & 70\% & 80\% & 51.6\% & 63.11\%  \\
    \hline
    Power  & - & 1.37 W & 0.33 W & 0.538 W & 2 W  \\
    \hline
    \end{tabular}}
\end{table}

\section{Conclusions}
\vspace*{-10pt}
This work has presented a novel astrocyte-augmented spiking neural network model implemented on CPU/GPU and FPGA platforms. The inclusion of astrocytes has shown significant improvements in the network's fault tolerance, demonstrating the potential benefits of astrocyte integration in artificial neural networks. Additionally, the use of FPGA hardware for this model leverages the advantages of parallel computation and on-the-fly hardware reconfiguration offered by DFX technology. The comparison with different computational architectures and previous works highlighted the strengths of our approach in terms of computational efficiency and network robustness. Future research in this direction could yield more sophisticated and efficient neuromorphic systems, thus paving the way for advanced applications in diverse areas such as robotics, bioinformatics, and cognitive computing.

\bibliographystyle{splncs04}
\bibliography{reference}

\end{document}